\documentclass[10pt,twocolumn,letterpaper]{article}

\usepackage[pagenumbers]{cvpr} % To force page numbers, e.g. for an arXiv version

\usepackage{verbatim} % 引入verbatim宏包

\usepackage{booktabs}
\usepackage{multirow}
\usepackage{makecell}
\usepackage{bm}
\usepackage{relsize}

\usepackage[pagebackref,breaklinks,colorlinks,allcolors=cvprblue]{hyperref}

\usepackage{multirow}
\usepackage{times}  % DO NOT CHANGE THIS
\usepackage{helvet}  % DO NOT CHANGE THIS
\usepackage{courier}  % DO NOT CHANGE THIS
\usepackage[hyphens]{url}  % DO NOT CHANGE THIS
\usepackage{graphicx} % DO NOT CHANGE THIS
\usepackage{natbib}  % DO NOT CHANGE THIS AND DO NOT ADD ANY OPTIONS TO IT
\usepackage{caption} % DO NOT CHANGE THIS AND DO NOT ADD ANY OPTIONS TO IT
\usepackage[utf8]{inputenc} % allow utf-8 input
\usepackage[T1]{fontenc}    % use 8-bit T1 fonts
\usepackage{url}            % simple URL typesetting
\usepackage{booktabs}       % professional-quality tables
\usepackage{amsfonts}       % blackboard math symbols
\usepackage{nicefrac}       % compact symbols for 1/2, etc.
\usepackage{microtype}      % microtypography
\usepackage{xcolor}         % colors
\usepackage{multirow}
\usepackage{graphicx}
\usepackage{tabularray}
\usepackage{float}
\usepackage{amsmath}
\usepackage{amssymb}
\usepackage{bm}
\usepackage{colortbl}
\usepackage{tabularray}
\usepackage{float}
\usepackage{bm}
\usepackage{threeparttable}
\usepackage[algo2e]{algorithm2e}
\usepackage[capitalize]{cleveref}

\usepackage{algorithm}
\usepackage{algorithmic}
\usepackage{bm}
\usepackage[accsupp]{axessibility}  % Improves PDF readability for those with disabilities.

\definecolor{1}{rgb}{0.96, 0.57, 0.58}
\definecolor{2}{rgb}{0.98, 0.78, 0.57}
\definecolor{3}{rgb}{1.0, 1.0, 0.56}

\usepackage{multirow}
\usepackage{booktabs}
\usepackage{subcaption}
\usepackage{amsmath,amsfonts,bm,amsthm,amssymb}
\usepackage{graphics,graphicx,caption,float,color}
\usepackage{appendix}

\usepackage{listings}
\usepackage{xspace}
 
\usepackage{hhline}
\usepackage{multicol}
\usepackage{booktabs} 
\usepackage{makecell} % Xhline

\makeatletter
\newcommand*\bigcdot{\mathpalette\bigcdot@{.5}}
\newcommand*\bigcdot@[2]{\mathbin{\vcenter{\hbox{\scalebox{#2}{$\m@th#1\bullet$}}}}}
\makeatother

\definecolor{MyDarkBlue}{rgb}{0,0.08,1}
\definecolor{MyDarkGreen}{rgb}{0.02,0.6,0.02}
\definecolor{MyDarkRed}{rgb}{0.8,0.02,0.02}
\definecolor{MyDarkOrange}{rgb}{0.40,0.2,0.02}
\definecolor{MyPurple}{RGB}{111,0,255}
\definecolor{MyRed}{rgb}{1.0,0.0,0.0}
\definecolor{MyGold}{rgb}{0.75,0.6,0.12}
\definecolor{MyDarkgray}{rgb}{0.66, 0.66, 0.66}

\definecolor{Image}{RGB}{93,59,155}
\definecolor{Video}{RGB}{105,145,60}

\definecolor{ice}{RGB}{84,146,214}
\definecolor{fire}{RGB}{217,129,63}
\definecolor{baselinecolor}{gray}{.9}
\definecolor{poscolor}{RGB}{212,17,89}
\definecolor{negcolor}{RGB}{26,133,255}
\definecolor{bestcolor}{RGB}{238, 255, 238}
\newcolumntype{x}[1]{>{\centering\arraybackslash}p{#1pt}}
\newcolumntype{y}[1]{>{\raggedright\arraybackslash}p{#1pt}}
\newcolumntype{z}[1]{>{\raggedleft\arraybackslash}p{#1pt}}
\definecolor{cvprblue}{rgb}{0.21,0.49,0.74}
\usepackage[pagebackref,breaklinks,colorlinks,allcolors=cvprblue]{hyperref}

\title{WorldWarp: Propagating 3D Geometry with Asynchronous Video Diffusion
}

\author{Hanyang Kong\textsuperscript{\rm 1} \quad  Xingyi Yang\textsuperscript{\rm 2}\footnotemark[1] \quad  Xiaoxu Zheng\textsuperscript{\rm 1}   \quad Xinchao Wang\textsuperscript{\rm 1}\footnotemark[1]\\
{\fontsize{10}{15}\selectfont \textsuperscript{\rm 1}National University of Singapore} \quad 
{\fontsize{10}{15}\selectfont \textsuperscript{\rm 2}The Hong Kong Polytechnic University}\\
{\tt\small hanyang.k@u.nus.edu, xingyi.yang@polyu.edu.hk, xinchao@nus.edu.sg}\\
{\small \href{https://hyokong.github.io/worldwarp-page/}{https://hyokong.github.io/worldwarp-page/}}
}

\begin{document}

\twocolumn[{
\renewcommand\twocolumn[1][]{#1}
\maketitle

\begin{center}
    % \vspace{-8mm}
    \setlength{\abovecaptionskip}{6pt}
    \centering
    \includegraphics[width=1.0\linewidth]{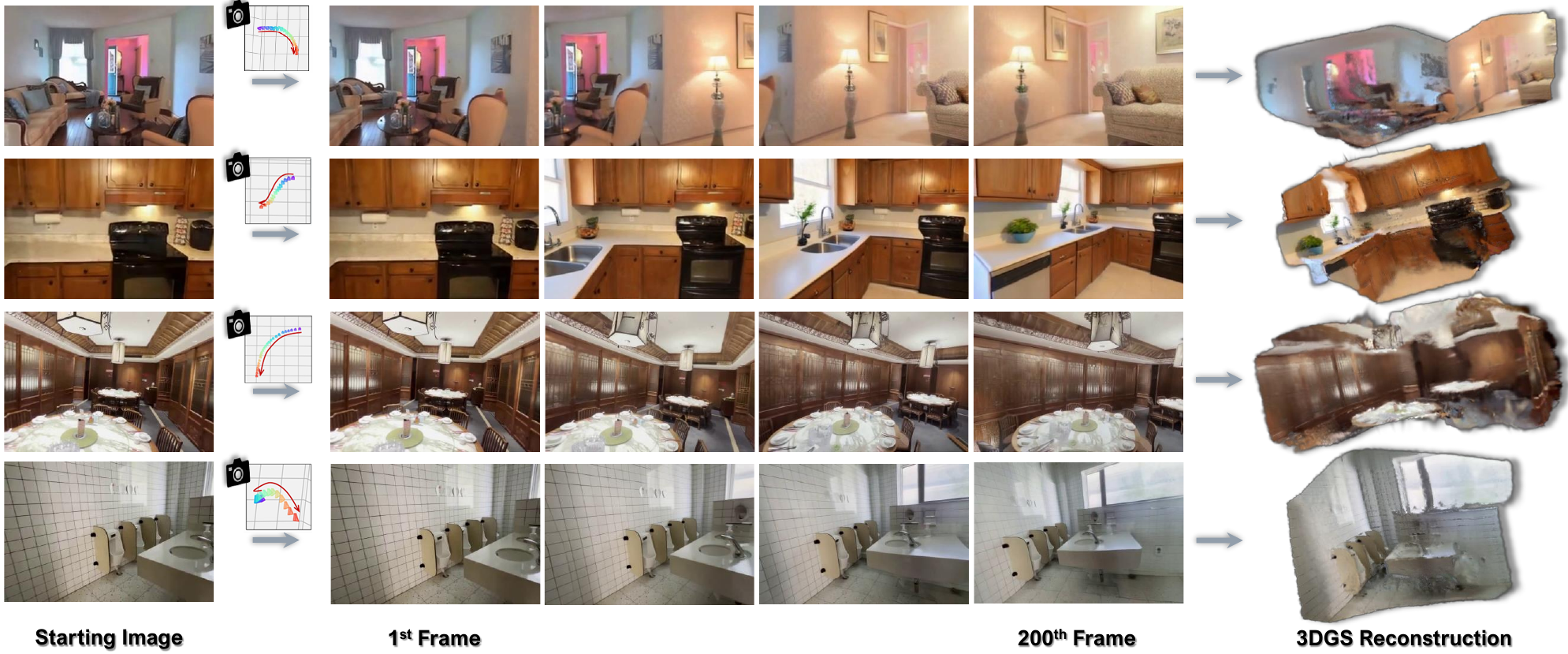}
    \captionof{figure}{\textbf{WorldWarp: Long-range novel view synthesis from a single image.} 
    Given only a single starting image (left) and a specified camera trajectory, our method generates a long and coherent video sequence. 
    The core of our approach is to generate the video chunk-by-chunk, where each new chunk is conditioned on forward-warped ``hints" from the previous one. 
    A novel diffusion model then generates the next chunk by correcting these hints and filling in occlusions using a spatio-temporal varying noise schedule. 
    The high geometric consistency of our 200-frame generated sequence is demonstrated by its successful reconstruction into a high-fidelity 3D Gaussian Splatting (3DGS)~\cite{3dgs} model (right). 
    This highlights our model's robust understanding of 3D geometry and its capability to maintain long-term consistency.} 
    \label{fig:teaser} 
\end{center}
}
\vspace{0.5cm}
]

\renewcommand{\thefootnote}{\fnsymbol{footnote}}
\footnotetext[1]{Corresponding author.}

% \maketitle
\begin{abstract}
Generating long-range, geometrically consistent video presents a fundamental dilemma: while consistency demands strict adherence to 3D geometry in pixel space, state-of-the-art generative models operate most effectively in a camera-conditioned latent space. This disconnect causes current methods to struggle with occluded areas and complex camera trajectories. To bridge this gap, we propose WorldWarp, a framework that couples a 3D structural anchor with a 2D generative refiner. To establish geometric grounding, WorldWarp maintains an online 3D geometric cache built via Gaussian Splatting (3DGS). By explicitly warping historical content into novel views, this cache acts as a structural scaffold, ensuring each new frame respects prior geometry. However, static warping inevitably leaves holes and artifacts due to occlusions. We address this using a Spatio-Temporal Diffusion (ST-Diff) model designed for a "fill-and-revise" objective. Our key innovation is a spatio-temporal varying noise schedule: blank regions receive full noise to trigger generation, while warped regions receive partial noise to enable refinement. By dynamically updating the 3D cache at every step, WorldWarp maintains consistency across video chunks. Consequently, it achieves state-of-the-art fidelity by ensuring that 3D logic guides structure while diffusion logic perfects texture. Project page: \href{https://hyokong.github.io/worldwarp-page/}{https://hyokong.github.io/worldwarp-page/}.

\end{abstract}    
\section{Introduction}
\label{sec:intro}

Novel View Synthesis (NVS) has emerged as a cornerstone problem in computer vision and graphics, with transformative applications in virtual reality, immersive telepresence, and generative content creation. While traditional NVS methods excel at view interpolation, which generates new views \textit{within} the span of existing camera poses~\cite{nerf,3dgs,mipnerf}, the frontier of the field lies in \textbf{view extrapolation}~\cite{infinitenature,infinitenature0,genwarp,cameractrl,motionctrl,vmem}. This far more challenging task involves generating long, continuous camera trajectories that extend significantly beyond the original scene, effectively synthesizing substantial new content and structure~\cite{infinitenature,genwarp}. The ultimate goal is to enable interactive exploration of dynamic, 3D-consistent worlds from only a limited set of starting images.

The central challenge in generating long-range, camera-conditioned video lies in finding an effective 3D conditioning. Existing works have largely followed two main strategies. \emph{The first} is camera pose encoding, which embeds abstract camera parameters as a latent condition~\cite{zero123,cameractrl,motionctrl,genwarp,zeronvs,kong2024eschernet}. This approach, however, relies heavily on the diversity of the training dataset and often fails to generalize to Out-Of-Distribution (OOD) camera poses, while also providing minimal information about the underlying 3D scene content~\cite{fridman2023scenescape,hollein2023text2room,muller2024multidiff,popov2025camctrl3d,viewcrafter,vmem}. \emph{The second strategy}, which uses an explicit 3D spatial prior, was introduced to solve this OOD issue~\cite{fridman2023scenescape,hollein2023text2room,viewcrafter,vmem}. While these priors provide robust geometric grounding, they are imperfect, suffering from occlusions (blank regions) and distortions from 3D estimation errors~\cite{genwarp,viewcrafter}. This strategy typically employs standard inpainting or video generation techniques~\cite{fridman2023scenescape,hollein2023text2room,vmem}, which are ill-suited to simultaneously handle the severe disocclusions and the geometric distortions present in the warped priors, leading to artifacts and inconsistent results.

To address this critical gap, we propose \textbf{WorldWarp}, a novel framework that generates long-range, geometrically-consistent novel view sequences. Our core insight is to break the strict causal chain of AR models and the static nature of explicit 3D priors. Instead, WorldWarp operates via an autoregressive inference pipeline that generates video \textit{chunk-by-chunk} (see \cref{fig:inference}). The key to our system is a \textbf{Spatio-Temporal Diffusion (ST-Diff)} model~\cite{wan,ldm}, which is trained with a powerful bidirectional, non-causal attention mechanism. This non-causal design is explicitly enabled by our core technical idea: using forward-warped images from future camera positions as a dense, explicit 2D spatial prior~\cite{ttt3r}. At each step, we build an "online 3D geometric cache" using 3DGS~\cite{3dgs}, which is optimized only on the most recent, high-fidelity generated history. This cache then renders high-quality warped priors for the \textit{next} chunk, providing ST-Diff with a rich, geometrically-grounded signal that guides the generation of new content and fills occlusions.

The primary advantage of WorldWarp is its ability to avoid the irreversible error propagation that plagues prior work~\cite{genwarp,viewcrafter}. By dynamically re-estimating a short-term 3DGS cache at each step, our method continuously grounds itself in the most recent, accurate geometry, ensuring high-fidelity consistency over extremely long camera paths. We demonstrate the effectiveness of our approach through extensive experiments on challenging real-world and synthetic datasets for long-sequence view extrapolation, achieving state-of-the-art performance in both geometric consistency and visual fidelity.

In summary, our main contributions are:
\begin{itemize}
    \item \textbf{WorldWarp}, a novel framework for long-range novel view extrapolation that generates video chunk-by-chunk using an autoregressive inference pipeline.
    \item \textbf{Spatio-Temporal Diffusion (ST-Diff)}, a non-causal diffusion model that leverages bidirectional attention conditioned on forward-warped images as a dense geometric prior.
    \item An \textbf{online 3D geometric cache} mechanism, which uses test-time optimized 3DGS~\cite{3dgs} to provide high-fidelity warped priors while preventing the irreversible error propagation of static 3D representations.
    \item State-of-the-art performance on challenging view extrapolation benchmarks, demonstrating significantly improved geometric consistency and image quality over existing methods.
\end{itemize}

\vspace{-0.2cm}
\section{Related Works}
\label{sec:related_works}
\paragraph{Novel view synthesis.}

Novel view synthesis (NVS) is a challenging problem that can be categorized into two aspects: view interpolation~\cite{nerf,3dgs,mipnerf,zipnerf,mipsplatting,fsgs,dngaussian,sparsenerf,regnerf,gsw,wildgs,sls,robustnerf,nerfonthego} and extrapolation~\cite{infinitenature, infinitenature0,genwarp,causnvs,vmem,cameractrl,motionctrl,geogpt,photonvs,seva}. View interpolation task aims to generate novel views within the distributions of the training views~\cite{nerf,3dgs,mipnerf,zipnerf,mipsplatting} even if the training views are sparse~\cite{sparsenerf,regnerf,dngaussian,fsgs} or the training views are captured in the wild with occlusions~\cite{nerfonthego,sls,robustnerf}. View extrapolation tasks~\cite{infinitenature, infinitenature0,genwarp,causnvs,vmem,cameractrl,motionctrl,geogpt,photonvs,seva} focus on generating novel views which are extended significantly beyond the original scenes, introducing substantial new contents, by leveraging powerful pre-trained video diffusion models~\cite{ldm,wan,opensora,cogvideo,cogvideox}.

\vspace{-0.5cm}
\paragraph{Auto-regressive video diffusion models.}

The field of video generation has seen a prominent trend towards either diffusion-based or autoregressive (AR) methodologies. Parallel (non-autoregressive) video diffusion systems often employ bidirectional attention to process and denoise all frames concurrently~\cite{ho2022video,blattmann2023align,Ho2022ImagenVH,videoworldsimulators2024,polyak2024movie,yang2025cogvideox,hacohen2024ltx,kong2024hunyuanvideo,blattmann2023stable,villegas2023phenaki,deng2024nova,gupta2024photorealistic,wang2025wan}. Conversely, AR-based techniques produce content in a sequential manner. This category encompasses several architectures, such as models based on pure next-token prediction~\cite{weissenborn2020scaling, kondratyuk2024videopoet,yan2021videogpt,wang2024loong,Bruce2024GenieGI,ren2025next,selfforcing}, more recent hybrid systems integrating AR and diffusion principles~\cite{weng2024art,chen2024diffusion,yin2025causvid,hu2024acdit,jin2024pyramidal,gu2025long,gao2024ca2,li2025arlon,liu2024mardini,zhang2025test}, and rolling diffusion variants that employ progressive noise schedules~\cite{ruhe2024rolling,kim2025fifo,xie2024progressive,zhang2025packing,magi1,sun2025ar}. However, these AR strategies are ill-suited for this work's specific task. Learning an effective camera embedding for them is non-trivial, and their causal structure is incompatible with using warped images from \textit{future camera positions} as conditional hints. Consequently, this work employs a non-autoregressive framework~\cite{dfot} to leverage this future information.

\vspace{-0.5cm}
\paragraph{Camera pose encoding and 3D explicit spatial priors.}

\begin{figure*}[ht]
    \centering 
    \includegraphics[width=0.99\textwidth]{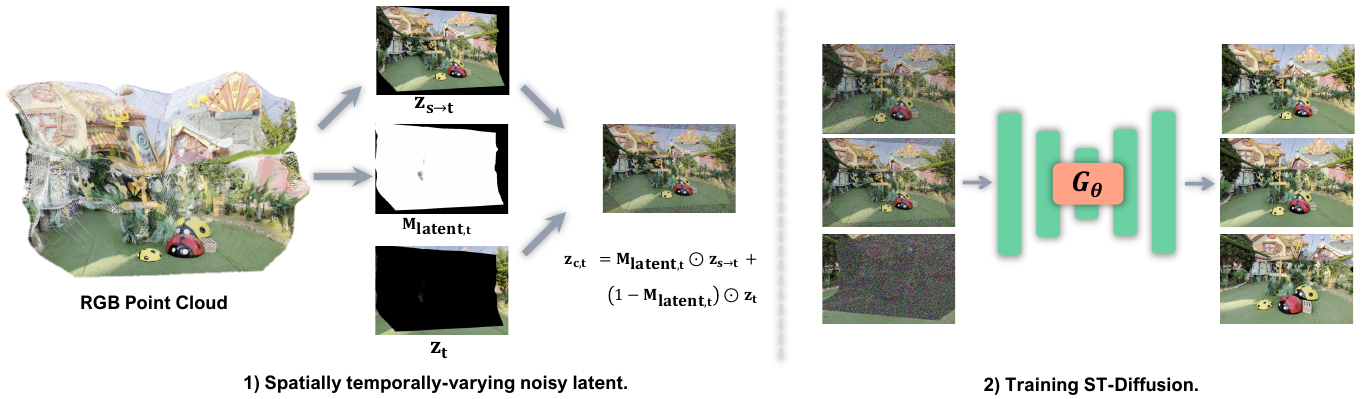}
      \captionsetup{font=small}
    \caption{
    \textbf{Training pipeline of our ST-Diff model.}
    \textbf{1) Spatially temporally-varying noisy latent:} The process begins by rendering a warped image and a validity mask from an RGB point cloud (images are shown for illustration, as operations are in latent space). The warped image is encoded to get \( \mathbf{z}_{s \to t} \), and the ground-truth image is encoded to get \( \mathbf{z}_t \). A "clean composite" latent \( \mathbf{z}_{c,t} \) is created by combining the valid warped regions from \( \mathbf{z}_{s \to t} \) with the blank regions from \( \mathbf{z}_t \), using the downsampled mask \( \mathbf{M}_{\text{latent}} \).
    \textbf{2) Training ST-Diffusion:} This composite latent sequence is noised according to our spatio-temporal schedule, resulting in a noisy latent sequence (visualized as a stack) where the noise level for each latent varies across different frames and spatial regions. The resulting noisy latents are fed into our model \( \bm{G}_\theta \), which is trained to predict the target velocity (defined as \( \mathbf{\epsilon}_t - \mathbf{z}_t \)), forcing it to learn the flow from the noisy composite latent back towards the original ground-truth latent sequence \( \mathcal{Z} \).
}
\label{fig:training_pipeline}
\end{figure*}

Spatially consistent view generation relies on conditioning. One method, camera pose encoding, models camera geometry using absolute extrinsics~\cite{zero123,cameractrl,motionctrl,genwarp,zeronvs} or relative representations like CaPE~\cite{kong2024eschernet}. While useful for viewpoint control, these encodings lack 3D scene content. An alternative, explicit 3D spatial priors, builds 3D models (\eg, meshes, point clouds, 3DGS~\cite{3dgs})~\cite{fridman2023scenescape, hollein2023text2room, muller2024multidiff, popov2025camctrl3d, viewcrafter,vmem} for re-projection and inpainting. This provides geometric grounding but suffers from error propagation from the initial 3D estimation~\cite{genwarp, viewcrafter} and high computational cost. Instead, we utilize forward-warped images from future camera positions as a distinct explicit prior. These warped images serve as a dense, geometrically-grounded 2D hint, bypassing the error-prone and costly 3D reconstruction pipeline while offering a richer conditional signal than mere pose encoding.

\vspace{-0.2cm}
\section{Preliminaries}
\label{sec:preliminaries}

\subsection{Camera-Conditioned Video Generation}

One major challenge in adding precise camera control to video diffusion models is finding a good way to represent 3D camera movement. Simply using raw camera intrinsics $\mathbf{K}$ and extrinsics $\mathbf{E}$ is often suboptimal, as their numerical values (e.g., translation $\mathbf{t}$) are unconstrained and difficult for a network to correlate with visual content.

A more effective paradigm is to translate these abstract parameters into a dense, pixel-wise representation that provides a clearer geometric interpretation. For example, Plücker embeddings~\cite{sitzmann2021light} define a 6D ray vector for each pixel. This transforms the abstract matrices into a dense tensor $\mathbf{P} \in \mathbb{R}^{n \times 6 \times h \times w}$, which is much more informative for the diffusion model. This principle of using dense, geometrically-grounded priors is a key consideration for enabling fine-grained camera control.

\subsection{Diffusion Forcing and Non-Causal Priors}

The Diffusion Forcing Transformer (DFoT)~\cite{dfot} paradigm re-frames the noising operation as progressive masking, where each frame $\mathbf{x}_t$ in a video is assigned an independent noise level $k_t \in [0, 1]$. This contrasts with conventional models that use a single noise level $k$ for all frames. The model $\bm{\hat\epsilon}_\theta$ is then trained on a per-frame noise prediction loss:
\begin{equation}
\label{eq:df_loss}
\mathcal{L} = \mathop{\mathlarger{\mathbb{E}}}_{\substack{k_{\mathcal{T}}, \mathcal{X}, \mathcal{E}}}
\Big[ \sum_{t=1}^T \left\| \bm{\hat\epsilon}_\theta(\mathcal{X}^k, k_{\mathcal{T}})_t - \mathbf{\epsilon}_t \right\|_2^2 \Big]
\end{equation}
The critical advantage of this per-frame noise approach is that it enables a model to be trained with non-causal attention, learning to denoise a frame by conditioning on an arbitrary, partially-masked set of other frames.

This non-causal paradigm is particularly well-suited for our task. In typical video generation, a causal architecture is necessary as the future is unknown. However, in camera-conditioned novel view synthesis, we can generate a strong, geometry-consistent \emph{prior} for all future frames simultaneously via forward-warping. These warped images provide a powerful non-causal conditioning signal. This insight is the foundation of our ST-Diff model, allowing us to discard restrictive causal constraints and employ a bidirectional, spatio-temporal diffusion strategy.

\section{Method}
\label{sec:method}

\subsection{Spatio-Temporal Diffusion with Warped Priors}
\label{ssec:method_warp}

We address the task of novel view synthesis, where the goal is to generate a target view $\mathbf{x}_t$ given a source view $\mathbf{x}_s$ and corresponding camera poses $\{\mathbf{p}_s, \mathbf{p}_t\}$. To this end, we introduce Spatio-Temporal Diffusion with Warped Priors (ST-Diff), a bidirectional diffusion model designed for this task. Unlike causal, autoregressive video generation, where future frames are unknown, the camera-conditioned setting allows us to form a strong geometric prior for the target frame by projecting the source view. This key insight allows us to discard causal constraints and employ a more powerful bidirectional attention mechanism across all frames.

Our method first prepares geometric priors in the \textbf{pixel space} and then performs all diffusion, compositing, and noising operations in the \textbf{latent space} using a pre-trained VAE encoder $\mathcal{E}(\cdot)$ and decoder $\mathcal{D}(\cdot)$~\cite{wan,ldm}. We use $\mathbf{x}$ to denote data in pixel-space and $\mathbf{z}$ for latent-space data.

% \vspace{-0.5cm}
\paragraph{One-to-all pixel-space warping.}
Given a training video sequence $\mathcal{X} = \{\mathbf{x}_i\}_{i=1}^T$, we first sample a single source frame $\mathbf{x}_s$ from the sequence. We then create a full sequence of warped priors by warping this single source frame $\mathbf{x}_s$ to every other frame's viewpoint, including its own. To do this, we use pre-estimated depth maps $\mathbf{D}_i$ and camera parameters (extrinsics $\mathbf{E}_i$ and intrinsics $\mathbf{K}_i$) for all frames, obtained from a 3D geometry foundation model~\cite{ttt3r}. First, the source image $\mathbf{x}_s$ and its depth $\mathbf{D}_s$ are unprojected into a 3D RGB point cloud $\mathcal{P}_s$:
\begin{equation}
\mathbf{p}_{\text{cam}}^{(u,v)} = \mathbf{D}_s(u, v) \cdot \mathbf{K}_s^{-1} [u, v, 1]^T
\end{equation}
\begin{equation}
\mathcal{P}_s = \{ (\mathbf{E}_s \mathbf{p}_{\text{cam}}^{(u,v)}, \mathbf{x}_s(u,v)) \}_{u,v}
\end{equation}
This single point cloud $\mathcal{P}_s$ is then rendered into all $T$ target viewpoints using a differentiable point-based renderer. This "one-to-all" warping process yields two new sequences: a \textbf{warped prior sequence}, $\mathcal{X}_{s \to \mathcal{V}} = \{\mathbf{x}_{s \to t}\}_{t=1}^T$, and a corresponding \textbf{validity mask sequence}, $\mathcal{M} = \{\mathbf{M}_t\}_{t=1}^T$. Each mask $\mathbf{M}_t$ indicates which pixels in $\mathbf{x}_{s \to t}$ were successfully rendered from $\mathcal{P}_s$.

% \vspace{-0.4cm}
\paragraph{Latent-space composite sequence.}
The training pipeline of our WorldWarp is illustrated in \cref{fig:training_pipeline}. With the pixel-space assets prepared, we move entirely to the latent space. We separately encode the new warped sequence $\mathcal{X}_{s \to \mathcal{V}}$ and the original ground-truth sequence $\mathcal{X}$. We encode both:
\begin{equation}
\mathcal{Z}_{s \to \mathcal{V}} = \{\mathcal{E}(\mathbf{x}_{s \to t})\}_{t=1}^T \quad \text{and} \quad \mathcal{Z} = \{\mathcal{E}(\mathbf{x}_t)\}_{t=1}^T.
\end{equation}
We also downsample the mask sequence $\mathcal{M}$ to match the latent dimensions, yielding $\mathcal{M}_{\text{latent}} = \{\mathbf{M}_{\text{latent}, t}\}_{t=1}^T$. A clean composite latent sequence $\mathcal{Z}_c$ is then created in the latent space. For each frame $t$, the composite $\mathbf{z}_{c,t}$ takes its features from the warped latent $\mathbf{z}_{s \to t}$ in valid ("warped") regions and fills the remaining ("filled") regions with features from the ground-truth latent $\mathbf{z}_t$ (which is the $t$-th element of $\mathcal{Z}$):
\begin{equation}
\mathbf{z}_{c,t} = \mathbf{M}_{\text{latent}, t} \odot \mathbf{z}_{s \to t} + (1 - \mathbf{M}_{\text{latent}, t}) \odot \mathbf{z}_t \quad \text{for } t=1...T
\end{equation}
This entire sequence $\mathcal{Z}_c = \{\mathbf{z}_{c,t}\}_{t=1}^T$ serves as the $x_0$-equivalent (clean signal) for the diffusion model.

% \vspace{-0.5cm}
\paragraph{Spatially and temporally-varying noise.}
Our noising strategy extends the per-frame independent noise concept with a new, region-specific dimension, as shown in \cref{fig:training_pipeline}. The noise applied is varied at two levels simultaneously. First, at a \textbf{temporal} level, each frame $t$ in the sequence $\mathcal{Z}_c$ gets a different, independently sampled noise schedule. Second, at a \textbf{spatial} level, we apply different noise levels *within* each frame, distinguishing between the "warped" and "filled" regions. For each frame $t$, we therefore sample a pair of noise levels, $(\sigma_{\text{warped}, t}, \sigma_{\text{filled}, t})$. A spatially-varying noise map $\mathbf{\Sigma}_t$ is constructed using the latent-space mask:
\begin{equation}
\mathbf{\Sigma}_t = \mathbf{M}_{\text{latent}, t} \odot \sigma_{\text{warped}, t} + (1 - \mathbf{M}_{\text{latent}, t}) \odot \sigma_{\text{filled}, t}
\end{equation}
We then generate the final noisy input sequence $\mathcal{Z}_{\text{noisy}} = \{\mathbf{z}_{\text{noisy}, t}\}_{t=1}^T$ by sampling a noise sequence $\mathcal{E} = \{\mathbf{\epsilon}_t\}_{t=1}^T \sim \mathcal{N}(0, \mathbf{I})$:
\begin{equation}
\mathbf{z}_{\text{noisy}, t} = (1 - \mathbf{\Sigma}_t) \odot \mathbf{z}_{c, t} + \mathbf{\Sigma}_t \odot \mathbf{\epsilon}_t
\end{equation}
A key architectural modification is required to process this spatiallyand  temporally varying noise. Standard diffusion models~\cite{wan} typically accept a single timestep embedding (e.g., shape $B \times 1$) for an entire image or video chunk. Our ST-Diff model, however, is adapted to process a unique noise level for every token. We broadcast the noise map sequence $\mathbf{\Sigma}_{\mathcal{V}} = \{\mathbf{\Sigma}_t\}_{t=1}^T$ to the full latent sequence dimensions ($B \times T \times H' \times W'$) and pass it through the time embedding network, thus generating a unique time-axis and spatial-axis embedding for each corresponding token.

\begin{figure*}[t]
    \centering 
    \includegraphics[width=0.99\textwidth]{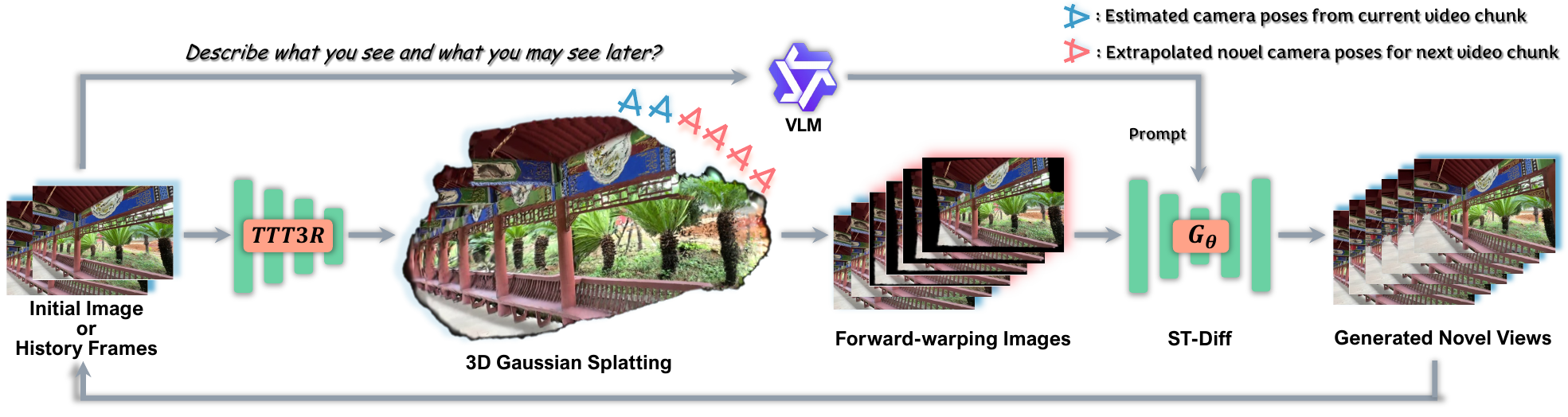}
      \captionsetup{font=small}
    \caption{\textbf{The autoregressive inference pipeline of WorldWarp.} 
    At each iteration $k$, the available history (either the initial images or the previously generated $k-1$ chunk) is processed. 
    First, TTT3R estimates camera poses and an initial 3D point cloud. 
    This geometry is used to optimize a 3D Gaussian Splatting (3DGS) representation, which serves as a high-fidelity 3D cache. 
    Concurrently, a VLM generates a descriptive text prompt, and novel camera poses are extrapolated for the next chunk. 
    The optimized 3DGS renders forward-warped images at these new poses. 
    These warped priors, along with the VLM prompt, are fed into our non-causal ST-Diff model ($G_\theta$) to denoise and generate the $k$-th chunk of novel views. 
    The process then repeats, using the newly generated chunk as the history for the next iteration.} 
    \label{fig:inference}
\end{figure*}

% \vspace{-0.4cm}
\paragraph{Training objective.}
We train our ST-Diff model $\bm{G}_\theta$ which takes the entire noisy sequence $\mathcal{Z}_{\text{noisy}}$, the sequence of noise maps $\mathbf{\Sigma}_{\mathcal{V}}$, and other conditioning $\mathbf{c}$ (e.g., text, camera poses) as input. Critically, the model is trained to denoise the composite sequence $\mathcal{Z}_{\text{noisy}}$ while regressing towards a target defined by the \emph{original ground-truth latent sequence} $\mathcal{Z}$. The target velocity sequence is $\mathcal{V}_{\text{target}} = \{\mathbf{\epsilon}_t - \mathbf{z}_t\}_{t=1}^T$. Our training objective is the $L_2$ loss, summed over the entire sequence:
\begin{equation}
\mathcal{L} = \mathbb{E}_{\mathcal{Z}, \mathcal{Z}_c, \mathcal{E}, \mathbf{\Sigma}_{\mathcal{V}}, \mathbf{c}} \left[ \sum_{t=1}^T \left\| \mathbf{v}_{\theta, t} - (\mathbf{\epsilon}_t - \mathbf{z}_t) \right\|_2^2 \right]
\end{equation}
where $\mathcal{V}_\theta = \{\mathbf{v}_{\theta, t}\}_{t=1}^T = \bm{G}_\theta(\mathcal{Z}_{\text{noisy}}, \mathbf{\Sigma}_{\mathcal{V}}, \mathbf{c})$. This loss forces the model to learn the complex relationship between the warped, GT-filled, and final target latents across the entire video.

\subsection{Autoregressive Inference Pipeline}
\label{ssec:method_inference}

The inference process is illustrated in \cref{fig:inference}. Our inference process generates novel view sequences autoregressively, producing a video chunk-by-chunk in a for-loop manner. Unlike training, which uses a fixed-radius point cloud representation, our inference pipeline leverages a dynamic, test-time optimized 3D representation as an explicit geometric cache. This process, illustrated in \cref{fig:inference}, integrates 3D Gaussian Splatting~\cite{3dgs} (3DGS) for high-fidelity warping and a Vision-Language Model (VLM)~\cite{qwen} for semantic guidance.
% \vspace{-0.3cm}
\paragraph{Online 3D Geometric Cache.}
At the beginning of each iteration $k$ of the generation loop, we take the available history (either the initial source views for $k=1$ or the video chunk generated in the previous iteration $k-1$). We first process these frames using a 3D geometry model (TTT3R)~\cite{ttt3r} to estimate their camera poses and an initial 3D point cloud. This point cloud is then used to initialize a 3D Gaussian Splatting (3DGS) representation, which we optimize for a few hundred steps (e.g., 200 steps) using the history frames and their estimated poses. This resulting online-optimized 3DGS model serves as an explicit, high-fidelity 3D representation cache. Compared to the fixed-radius point clouds used during training, this 3DGS provides significantly higher-quality features for the non-blank (warped) regions, which is critical for maintaining geometric consistency.

\begin{figure*}[t]
    \centering 
    \includegraphics[width=0.99\textwidth]{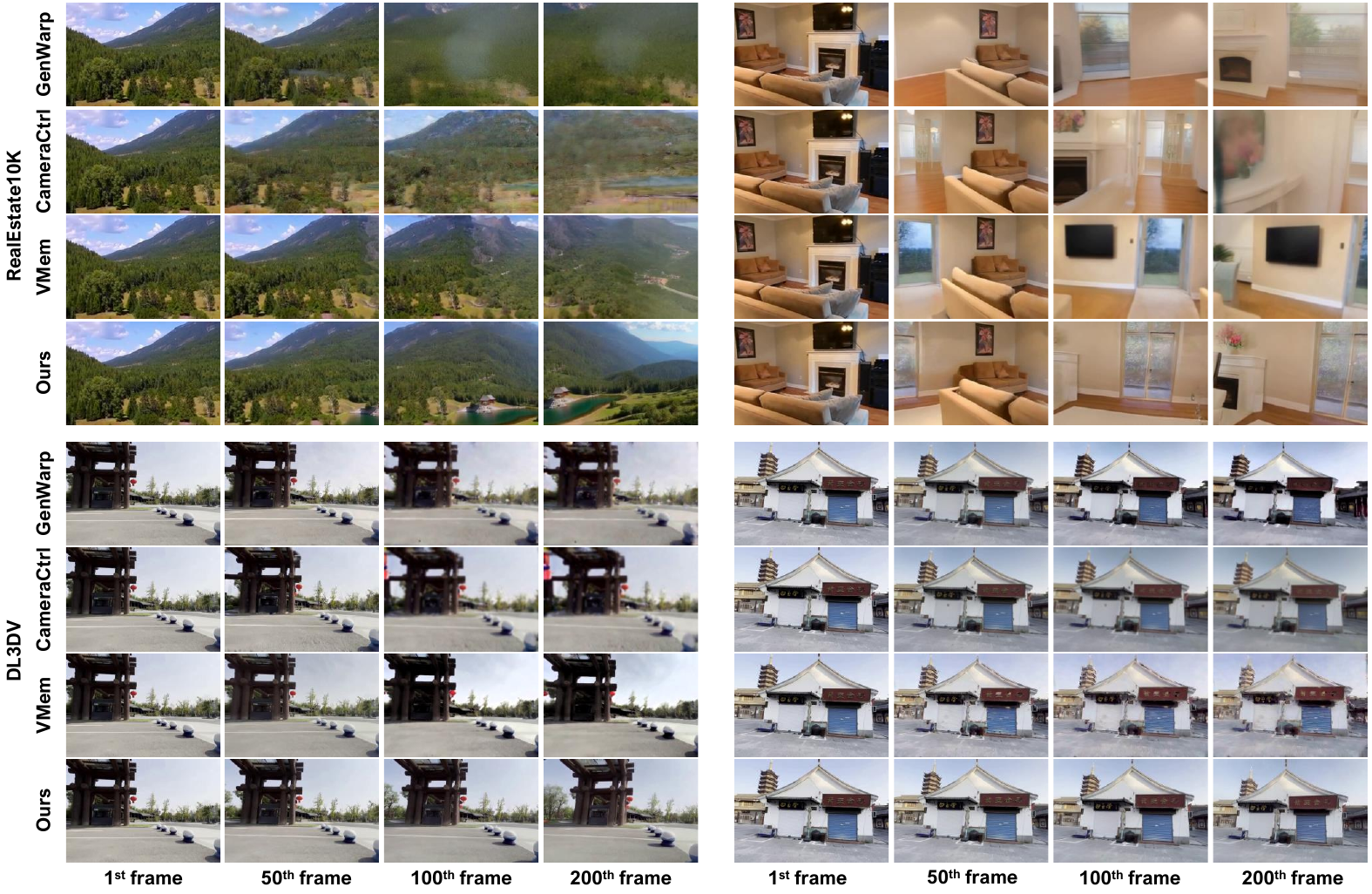}
      \captionsetup{font=small}
      \vspace{-0.3cm}
    \caption{\textbf{Qualitative comparisons on the RealEstate10K~\cite{re10k} and DL3DV~\cite{dl3dv} datasets.} 
    We visualize videos generated by our method against those by GenWarp~\cite{genwarp}, CameraCtrl~\cite{cameractrl}, and VMem~\cite{vmem}. Our WorldWarp generalizes to diverse camera motion, showcasing the spatial and temporal consistency.} 
    \label{fig:result}
\end{figure*}

% \vspace{-0.5cm}
\paragraph{Chunk-based Generation with ST-Diff.}
With the geometric and semantic conditioning prepared, we first render the sequence of prior images, $\mathcal{X}_{s \to \mathcal{V}}$, from the 3DGS cache. These are encoded into latents $\mathcal{Z}_{s \to \mathcal{V}} = \{\mathbf{z}_{s \to t}\}_{t=1}^T$, and we also obtain the corresponding latent-space masks $\mathcal{M}_{\text{latent}} = \{\mathbf{M}_{\text{latent}, t}\}_{t=1}^T$. Our goal is twofold: to \textbf{fill in} the blank (occluded) regions and to \textbf{revise} the non-blank (warped) regions, which may suffer from blur or distortion.

We achieve this by initializing the reverse diffusion process from a spatially-varying noise level, analogous to image-to-video translation. Let the full reverse schedule consist of $N$ timesteps, from $T_N=1000$ down to $T_1=1$. We define a strength parameter $\tau \in [0, 1]$, which maps to an intermediate timestep $T_{\text{start}}$ and its corresponding noise level $\sigma_{\text{start}}$. We set the noise level for the \textbf{blank} (filled) regions to $\sigma_{\text{filled}} = \sigma_{T_N}$, which corresponds to pure noise.

For each frame $t$, we construct a spatially-varying noise map $\mathbf{\Sigma}_{\text{start}, t}$ using the latent-space mask:
\begin{equation}
\mathbf{\Sigma}_{\text{start}, t} = \mathbf{M}_{\text{latent}, t} \odot \sigma_{\text{start}} + (1 - \mathbf{M}_{\text{latent}, t}) \odot \sigma_{\text{filled}}
\end{equation}
We then generate the initial noisy latent sequence $\mathcal{Z}_{\text{start}} = \{\mathbf{z}_{\text{start}, t}\}_{t=1}^T$ for the reverse process. This is done by applying the noise map $\mathbf{\Sigma}_{\text{start}, t}$ to the warped latent $\mathbf{z}_{s \to t}$, using a sampled Gaussian noise $\mathbf{\epsilon}_t$:
\begin{equation}
\mathbf{z}_{\text{start}, t} = (1 - \mathbf{\Sigma}_{\text{start}, t}) \odot \mathbf{z}_{s \to t} + \mathbf{\Sigma}_{\text{start}, t} \odot \mathbf{\epsilon}_t
\end{equation}
This formulation effectively initializes the blank regions with pure noise (as $\sigma_{\text{filled}} \approx 1.0$) while applying a partial, strength-controlled noising to the warped regions.

Our ST-Diff model ($\bm{G}_\theta$) then takes this spatially-mixed latent sequence $\mathcal{Z}_{\text{start}}$, the VLM text prompt, and the corresponding spatially-varying time embeddings as input. It denoises the sequence beginning from its spatially-varying timesteps (e.g., $T_{\text{start}}$ for warped regions and $T_N$ for blank regions) down to $T_1$ to generate the $k$-th chunk of novel views. This newly generated chunk is then used as the history for the next iteration ($k+1$), and the entire process repeats.
\section{Experiments}
\label{sec:experiments}

\subsection{Implementation Details.}

We fine-tune WorldWarp based on Wan2.1-T2V-1.3B~\cite{wan} model, with resolution 720x480 and batch size 8, on 8 H200 GPUs for 10k iterations.
We apply TTT3R~\cite{ttt3r} as the 3D reconstruction foundation model for estimating camera parameters and depth maps. Please refer to the supplementary material for more details.

\paragraph{Datasets and evaluation metrics.} We conduct experiments on two public scene-level datasets: RealEstate10K (Re10K)~\cite{re10k} and DL3DV~\cite{dl3dv} datasets.  Our evaluation of novel view synthesis quality comprises three main components: 1) Perceptual quality: We measure the distributional similarity between generated views and the test set using the Fréchet Image Distance (FID)~\cite{fid}. 2) Detail preservation: Following \cite{lookout}, we assess the model's ability to preserve image details across views by computing PSNR, SSIM~\cite{ssim}, and LPIPS~\cite{lpips}. 3) Geometric alignment: We evaluate camera pose accuracy against the ground truth ($\mathbf{R}_{\text{gt}}$, $\mathbf{t}_{\text{gt}}$), following \cite{motionctrl}. We use DUST3R~\cite{dust3r} to extract poses ($\mathbf{R}_{\text{gen}}$, $\mathbf{t}_{\text{gen}}$) from generated views. We then compute the rotation distance ($R_{\text{dist}}$) and translation distance ($t_{\text{dist}}$):
\begin{align*}
    R_{\text{dist}} &= \arccos\left\{ 0.5 (\text{tr}(\mathbf{R}_{\text{gen}}\mathbf{R}_{\text{gt}}^T) - 1)\right\} \\
    t_{\text{dist}} &= \|\mathbf{t}_{\text{gt}} - \mathbf{t}_{\text{gen}}\|_2,
\end{align*}
where $\text{tr}$ stands for the trace of a matrix. Per \cite{cameractrl}, estimated poses are expressed relative to the first frame, and translation is normalized by the furthest frame.

\begin{table*}
    [htbp]
    \centering
    \small
    \renewcommand{\arraystretch}{1.1}
    \caption{\textbf{Quantitative comparison for single-view NVS on the
    RealEstate10K~\cite{re10k} dataset.} We report performance for both \textbf{short-term}
    ($50^{th}$ frame) and \textbf{long-term} ($200^{th}$ frame) synthesis. For
    each metric, the \colorbox{1}{best}, \colorbox{2}{second best}, and \colorbox{3}{third
    best} results are highlighted. Our method significantly outperforms all
    baselines across most metrics, demonstrating superior quality and temporal consistency.
    }
    \vspace{-0.3cm}
    \scalebox{0.85}{
    \begin{tabular}{lcccccc|cccccc}
        \toprule \multicolumn{1}{c}{}                              & \multicolumn{6}{c}{\textbf{Short-term} ($50^{th}\text{frame}$)} & \multicolumn{6}{c}{\textbf{Long-term} ($200^{th}\text{frame}$)} \\
        \cmidrule(lr){2-7}\cmidrule(lr){8-13} \multicolumn{1}{c}{} & \textbf{PSNR} $\uparrow$                                        & \textbf{SSIM} $\uparrow$                                       & \textbf{LPIPS} $\downarrow$ & \textbf{FID} $\downarrow$ & $R_{\text{dist}}$ $\downarrow$ & $T_{\text{dist}}$ $\downarrow$ & \textbf{PSNR} $\uparrow$ & \textbf{SSIM} $\uparrow$ & \textbf{LPIPS} $\downarrow$ & \textbf{FID} $\downarrow$ & $R_{\text{dist}}$ $\downarrow$ & $T_{\text{dist}}$ $\downarrow$ \\
        \midrule InfiniteNature~\cite{infinitenature}              & 14.12                                                           & 0.192                                                          & 0.428                       & 27.35                     & 0.988                          & 0.521                          & 10.07                    & 0.109                    & 0.658                       & 46.12                     & 1.234                          & 0.902                          \\
        InfiniteNature-Zero~\cite{infinitenature0}                 & 14.31                                                           & 0.201                                                          & 0.409                       & 26.98                     & 0.931                          & 0.502                          & 10.22                    & 0.117                    & 0.647                       & 45.71                     & 1.201                          & 0.885                          \\
        GeoGPT~\cite{geogpt}                                       & 13.54                                                           & 0.186                                                          & 0.437                       & 26.52                     & 0.732                          & 0.413                          & 9.67                     & 0.089                    & 0.664                       & 41.52                     & \colorbox{2}{1.112}            & 0.876                          \\
        Lookout~\cite{lookout}                                     & 14.63                                                           & 0.216                                                          & 0.372                       & 28.67                     & 1.221                          & 0.864                          & 10.73                    & 0.163                    & 0.647                       & 57.82                     & 1.331                          & 0.912                          \\
        PhotoNVS~\cite{photonvs}                                   & 15.76                                                           & 0.247                                                          & 0.324                       & 21.62                     & 0.735                          & 0.464                          & 11.76                    & 0.176                    & 0.573                       & 41.67                     & 1.273                          & 0.628                          \\
        GenWarp~\cite{genwarp}                                     & 13.21                                                           & 0.252                                                          & 0.428                       & 29.51                     & 0.553                          & \colorbox{3}{0.059}            & 9.72                     & 0.192                    & 0.601                       & 36.12                     & 1.136                          & \colorbox{2}{0.446}            \\
        MotionCtrl~\cite{motionctrl}                               & 14.14                                                           & 0.258                                                          & 0.327                       & 19.12                     & 0.336                          & 0.353                          & 9.26                     & 0.187                    & 0.593                       & 35.21                     & 1.134                          & 0.697                          \\
        CameraCtrl~\cite{cameractrl}                               & 14.97                                                           & 0.271                                                          & 0.311                       & 20.07                     & 0.308                          & 0.267                          & 11.16                    & 0.183                    & 0.584                       & 35.07                     & 1.206                          & 0.704                          \\
        ViewCrafter~\cite{viewcrafter}                             & 17.23                                                           & 0.279                                                          & 0.367                       & 22.21                     & 1.242                          & 0.201                          & 9.96                     & 0.157                    & 0.578                       & 33.82                     & 1.571                          & 0.814                          \\
        SEVA~\cite{seva}                                           & \colorbox{2}{18.67}                                             & 0.394                                                          & 0.281                       & \colorbox{3}{17.14}       & \colorbox{3}{0.259}            & 0.116                          & \colorbox{3}{13.24}      & 0.227                    & \colorbox{3}{0.443}         & 28.47                     & \colorbox{3}{1.112}            & 0.731                          \\
        VMem~\cite{vmem}                                           & 18.19                                                           & \colorbox{3}{0.403}                                            & \colorbox{3}{0.273}         & \colorbox{2}{16.97}       & \colorbox{2}{0.221}            & \colorbox{2}{0.043}            & 14.91                    & \colorbox{3}{0.223}      & 0.471                       & \colorbox{3}{25.17}       & 1.132                          & \colorbox{3}{0.494}            \\
        DFoT~\cite{dfot}                                           & \colorbox{3}{18.53}                                             & \colorbox{2}{0.439}                                            & \colorbox{2}{0.265}         & 17.27                     & 0.326                          & 0.318                          & \colorbox{2}{15.21}      & \colorbox{2}{0.245}      & \colorbox{2}{0.418}         & \colorbox{2}{24.85}       & 1.643                          & 0.835                          \\
        \midrule

Ours                                             & \colorbox{1}{20.32}                                             & \colorbox{1}{0.527}                                            & \colorbox{1}{0.216}         & \colorbox{1}{15.56}       & \colorbox{1}{0.188}            & \colorbox{1}{0.039}            & \colorbox{1}{17.13}      & \colorbox{1}{0.281}      & \colorbox{1}{0.352}         & \colorbox{1}{21.75}       & \colorbox{1}{0.697}            & \colorbox{1}{0.203}            \\
        \bottomrule
    \end{tabular}
    }
    % \vspace{-0.2cm}
    
    \label{tab:re10k}
    % \vspace{-1mm}
\end{table*}

\subsection{Comparisons on the RealEstate10K Dataset}
% \vspace{-0.2cm}
We present a comprehensive quantitative evaluation on the RealEstate10K dataset in Table~\ref{tab:re10k}, assessing generation quality (PSNR, LPIPS) and camera pose accuracy ($R_{\text{dist}}$, $T_{\text{dist}}$) for short-term ($50^{th}$ frame) and long-term ($200^{th}$ frame) synthesis. Our method achieves state-of-the-art results, outperforming all baselines across all 12 metrics. This advantage is most pronounced in the challenging long-term setting: while most methods suffer significant quality degradation, our model maintains the highest PSNR (17.13) and LPIPS (0.352), surpassing strong competitors like SEVA, VMem, and DFoT. This high fidelity is crucial, as pose estimation (using Master3R) fails on the low-quality or blurry outputs from baselines. Consequently, our model achieves the lowest long-term pose error ($R_{\text{dist}}$ 0.697, $T_{\text{dist}}$ 0.203). This highlights a clear distinction: camera-embedding methods (MotionCtrl, CameraCtrl) suffer severe pose drift, and while 3D-aware methods (GenWarp, VMem) are more stable, our spatial-temporal noise diffusion strategy significantly surpasses both, proving its superior ability to mitigate cumulative camera drift. Qualitative results are in \cref{fig:result} and the supplementary.

\subsection{Comparisons on the DL3DV Dataset}

We further validate our model on the more challenging DL3DV dataset in \cref{tab:dl3dv}. Despite the complex trajectories degrading performance for all methods, our model maintains a commanding lead in all 12 metrics, demonstrating superior robustness. In the demanding long-term ($200^{th}$ frame) setting, our model's PSNR (14.53) decisively outperforms the next-best competitors, DFoT (13.51) and VMem (12.28). This fidelity is again proven critical for pose accuracy. On this complex dataset, our model remains the most stable, achieving the lowest $R_{\text{dist}}$ (1.007) and $T_{\text{dist}}$ (0.412). The weaknesses of competing approaches are magnified here, as 3D-aware methods like GenWarp ($1.351 R_{\text{dist}}$) and VMem ($1.419 R_{\text{dist}}$) lose stability. This proves our spatial-temporal noise diffusion strategy is more effective at preserving 3D consistency and mitigating severe camera drift on complex, long-range trajectories. Visualizations are in \cref{fig:result} and the supplementary material.

\begin{table*}
    [t]
    \centering
    \small
    \renewcommand{\arraystretch}{1.1}
    \caption{\textbf{Single-view NVS on DL3DV dataset~\cite{dl3dv}} Short-term evaluation
    is on the $50^{th}$ frame, and long-term is on frames $200^{th}$. This dataset
    is significantly more challenging due to complex camera trajectories and diverse
    environments. All methods show a noticeable performance drop compared to
    RealEstate10K~\cite{re10k}. For each metric, the \colorbox{1}{best}, \colorbox{2}{second
    best}, and \colorbox{3}{third best} results are highlighted. }%
    \vspace{-0.3cm}
    \scalebox{0.85}{
    \begin{tabular}{lcccccc|cccccc}
        \toprule \multicolumn{1}{c}{}                                             & \multicolumn{6}{c}{\textbf{Short-term} ($50^{th}\text{frame}$)} & \multicolumn{6}{c}{\textbf{Long-term} ($200^{th}\text{frame}$)} \\
        \cmidrule(lr){2-7}\cmidrule(lr){8-13} \multicolumn{1}{c}{\textbf{Method}} & \textbf{PSNR} $\uparrow$                                        & \textbf{SSIM} $\uparrow$                                       & \textbf{LPIPS} $\downarrow$ & \textbf{FID} $\downarrow$ & $R_{\text{dist}}$ $\downarrow$ & $T_{\text{dist}}$ $\downarrow$ & \textbf{PSNR} $\uparrow$ & \textbf{SSIM} $\uparrow$ & \textbf{LPIPS} $\downarrow$ & \textbf{FID} $\downarrow$ & $R_{\text{dist}}$ $\downarrow$ & $T_{\text{dist}}$ $\downarrow$ \\
        \midrule InfiniteNature~\cite{infinitenature}                             & 10.05                                                           & 0.112                                                          & 0.662                       & 51.45                     & 1.478                          & 1.131                          & 8.98                     & 0.100                    & 0.695                       & 54.32                     & 1.561                          & 1.522                          \\
        InfiniteNature-Zero~\cite{infinitenature0}                                & 10.21                                                           & 0.121                                                          & 0.648                       & 51.05                     & 1.432                          & 1.109                          & 9.12                     & 0.107                    & 0.685                       & 53.95                     & 1.528                          & 1.501                          \\
        GeoGPT~\cite{geogpt}                                                      & 9.83                                                            & 0.096                                                          & 0.688                       & 50.12                     & 1.553                          & 1.407                          & 8.52                     & 0.081                    & 0.773                       & 55.24                     & 1.851                          & 1.703                          \\
        Lookout~\cite{lookout}                                                    & 11.14                                                           & 0.131                                                          & 0.609                       & 69.53                     & 1.252                          & 1.058                          & 9.91                     & 0.117                    & 0.678                       & 75.06                     & \colorbox{3}{1.354}            & 1.552                          \\
        PhotoNVS~\cite{photonvs}                                                  & 12.02                                                           & 0.147                                                          & 0.558                       & 48.03                     & 1.404                          & 1.306                          & 10.83                    & 0.132                    & 0.609                       & 52.51                     & 1.708                          & 1.602                          \\
        GenWarp~\cite{genwarp}                                                    & 12.87                                                           & 0.201                                                          & 0.677                       & 44.04                     & 0.952                          & \colorbox{3}{0.381}            & 8.63                     & 0.092                    & 0.749                       & 48.13                     & \colorbox{2}{1.351}            & 0.953                          \\
        MotionCtrl~\cite{motionctrl}                                              & 13.34                                                           & 0.192                                                          & 0.698                       & 43.11                     & \colorbox{3}{0.863}            & 0.724                          & 8.12                     & 0.087                    & 0.779                       & 47.54                     & 1.452                          & 1.161                          \\
        CameraCtrl~\cite{cameractrl}                                              & 13.62                                                           & 0.212                                                          & 0.573                       & 32.53                     & 0.921                          & 0.832                          & 10.24                    & 0.127                    & 0.623                       & 46.92                     & 1.523                          & \colorbox{3}{0.924}            \\
        ViewCrafter~\cite{viewcrafter}                                            & 16.17                                                           & 0.226                                                          & 0.598                       & \colorbox{2}{31.02}       & 1.304                          & 0.953                          & 8.97                     & 0.112                    & 0.649                       & 45.23                     & 1.651                          & 1.052                          \\
        SEVA~\cite{seva}                                                          & 16.63                                                           & 0.331                                                          & \colorbox{3}{0.469}         & \colorbox{3}{31.04}       & 1.203                          & 0.851                          & 12.16                    & 0.181                    & \colorbox{3}{0.508}         & 36.03                     & 1.422                          & 0.954                          \\
        VMem~\cite{vmem}                                                          & \colorbox{2}{16.98}                                             & \colorbox{2}{0.348}                                            & \colorbox{2}{0.458}         & 31.52                     & \colorbox{2}{0.854}            & \colorbox{2}{0.352}            & 12.28                    & \colorbox{3}{0.197}      & \colorbox{2}{0.502}         & \colorbox{3}{35.52}       & 1.419                          & \colorbox{2}{0.858}            \\
        DFoT~\cite{dfot}                                                          & \colorbox{3}{16.13}                                             & \colorbox{3}{0.372}                                            & \colorbox{2}{0.402}         & 32.76                     & 1.139                          & 0.570                          & \colorbox{2}{13.51}      & \colorbox{2}{0.233}      & \colorbox{2}{0.471}         & \colorbox{2}{33.58}       & 1.685                          & 1.144                          \\
        \midrule

Ours                                                            & \colorbox{1}{18.10}                                             & \colorbox{1}{0.432}                                            & \colorbox{1}{0.315}         & \colorbox{1}{28.03}       & \colorbox{1}{0.433}            & \colorbox{1}{0.086}            & \colorbox{1}{14.53}      & \colorbox{1}{0.241}      & \colorbox{1}{0.413}         & \colorbox{1}{29.21}       & \colorbox{1}{1.007}            & \colorbox{1}{0.412}            \\
        \bottomrule
    \end{tabular}
    }
    
    \label{tab:dl3dv}
    \vspace{-1mm}
\end{table*}

\begin{figure}[htbp]
    \centering 
    \includegraphics[width=0.47\textwidth]{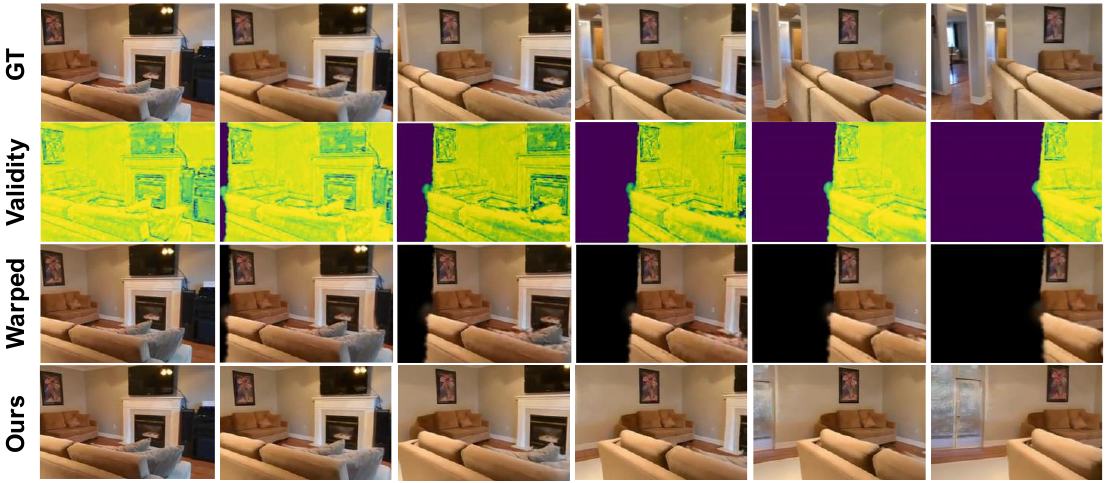}
      \captionsetup{font=small}
      \vspace{-0.3cm}
    \caption{\textbf{Illustration of the ST-Diff's generating process.} 
    We illustrate the GT images, the warped images which serve as the condition for ST-Diff, the corresponding validity mask, and our final generated frames. 
    The comparisons show that our ST-Diff successfully fills in the blank areas (initialized from a full noise level) while simultaneously revising distortions and enhancing details in the non-blank regions (initialized from a partial noise level) during the diffusion process.} 
    \label{fig:warp}
\end{figure}

\subsection{Ablation Study}
\vspace{-0.1cm}

\begin{table*}
    [t]
    \centering
    \small
    \renewcommand{\arraystretch}{1.1}
    \caption{\textbf{Ablation studies on the RealEstate10K~\cite{re10k} dataset.}
    We analyze the impact of our caching mechanism (top) and the spatial-temporal
    noise design (bottom). }
    \vspace{-0.3cm}
    \scalebox{0.8}{
    \begin{tabular}{lcccccc|cccccc}
        \toprule \multicolumn{1}{c}{}                              & \multicolumn{6}{c}{\textbf{Short-term} ($50^{th}\text{frame}$)} & \multicolumn{6}{c}{\textbf{Long-term} ($200^{th}\text{frame}$)} \\
        \cmidrule(lr){2-7}\cmidrule(lr){8-13} \multicolumn{1}{c}{} & \textbf{PSNR} $\uparrow$                                        & \textbf{SSIM} $\uparrow$                                       & \textbf{LPIPS} $\downarrow$ & \textbf{FID} $\downarrow$ & $R_{\text{dist}}$ $\downarrow$ & $T_{\text{dist}}$ $\downarrow$ & \textbf{PSNR} $\uparrow$ & \textbf{SSIM} $\uparrow$ & \textbf{LPIPS} $\downarrow$ & \textbf{FID} $\downarrow$ & $R_{\text{dist}}$ $\downarrow$ & $T_{\text{dist}}$ $\downarrow$ \\
        \midrule No Cache                                          & 14.19                                                           & 0.255                                                          & 0.331                       & 19.24                     & ---                            & ---                            & 9.22                     & 0.175                    & 0.598                       & 35.30                     & ---                            & ---                            \\
        Caching by RGB point cloud                                 & 15.12                                                           & 0.374                                                          & 0.269                       & 16.95                     & 0.192                          & 0.045                          & 11.12                    & 0.245                    & 0.412                       & 28.98                     & 0.703                          & 0.252                          \\
        Caching by online optimized 3DGS                           & 20.32                                                           & 0.527                                                          & 0.216                       & 15.56                     & 0.188                          & 0.039                          & 17.13                    & 0.281                    & 0.352                       & 21.75                     & 0.697                          & 0.203                          \\
        \midrule Full sequence noise                               & 17.08                                                           & 0.282                                                          & 0.364                       & 22.31                     & 1.235                          & 0.208                          & 9.92                     & 0.160                    & 0.580                       & 33.89                     & 1.574                          & 0.817                          \\
        Spatial-varying noise                                      & 18.23                                                           & 0.375                                                          & 0.305                       & 18.91                     & 0.232                          & 0.094                          & 13.95                    & 0.210                    & 0.492                       & 31.12                     & 1.040                          & 0.595                          \\
        Temporal-varying noise                                     & 18.09                                                           & 0.317                                                          & 0.298                       & 19.75                     & 0.258                          & 0.112                          & 13.20                    & 0.196                    & 0.513                       & 32.01                     & 1.209                          & 0.701                          \\
        Spatial-temporal-varying noise                             & 20.32                                                           & 0.527                                                          & 0.216                       & 15.56                     & 0.188                          & 0.039                          & 17.13                    & 0.281                    & 0.352                       & 21.75                     & 0.697                          & 0.203                          \\
        \bottomrule
    \end{tabular}
    }
    
    \label{tab:ablation}
    \vspace{-1mm}
\end{table*}

We conduct ablation studies on the RealEstate10K dataset in Table~\ref{tab:ablation} to validate our two core design choices: the 3DGS-based cache and the spatial-temporal noise diffusion model.

\noindent\textbf{Caching Mechanism.} We first analyze the effect of our caching module. The "No Cache" baseline, which relies only on the initial image, fails completely in long-term generation, with PSNR dropping to 9.22. This confirms the necessity of a 3D cache for long-range synthesis. We then compare our full model, "Caching by online optimized 3DGS," against "Caching by RGB point cloud." Although our model is trained on warped point clouds (with unoptimized, uniform radii), using a simple point cloud cache at inference ("Caching by RGB point cloud") yields significantly lower performance (11.12 PSNR) than our full model (17.13 PSNR). This demonstrates that using an online optimized 3DGS as the cache provides a much more robust and high-fidelity 3D representation. Notably, this 3DGS optimization is highly efficient, requiring only 500 steps per chunk. This result confirms that despite the modality gap between training (point clouds) and inference (3DGS), the superior representation quality of 3DGS leads to a substantial improvement in both generation quality and pose accuracy.

\noindent\textbf{Noise Diffusion Model.} The bottom half of the table validates our spatial-temporal noise diffusion design. Using a "Full sequence noise" (i.e., a standard video diffusion model) results in poor generation quality (9.92 long-term PSNR) and, critically, a catastrophic loss of camera control (1.574 $R_{\text{dist}}$). When using only "Spatial-varying noise," we observe a dramatic improvement in camera accuracy ($R_{\text{dist}}$ improves from 1.574 to 1.040), confirming that spatial noise is key for precise camera conditioning. Conversely, using only "Temporal-varying noise" improves generation quality (13.20 long-term PSNR) but fails to control the camera (1.209 $R_{\text{dist}}$). Our full "Spatial-temporal-varying noise" model successfully combines both benefits, achieving the best generation quality (17.13 PSNR) and the best camera accuracy (0.697 $R_{\text{dist}}$), demonstrating the necessity and efficacy of our proposed noise diffusion strategy.

\begin{table}[htbp]
\centering
\caption{Breakdown of latency and model size for each component in our pipeline. Times are in seconds (s).}
\vspace{-0.2cm}
\label{tab:inference_breakdown}
\resizebox{0.47\textwidth}{!}{
\begin{tabular}{l c c c c c c}
\toprule
& VLM Prompting & \shortstack{Estimating 3D \\ (TTT3R)} & \shortstack{Optimizing \\ 3DGS} & \shortstack{Forward \\ warping} &\shortstack{ST-Diff \\ 50 steps} & Total \\
\midrule 
\shortstack{Inference \\ time (s) } & 3.5 & 5.8 & 2.5 & 0.2 & 42.5 & 54.5 \\
\bottomrule 
\end{tabular}
}
\end{table}

\noindent\textbf{Inferencing efficiency.} We provide a detailed breakdown of the inference latency per video chunk in Table~\ref{tab:inference_breakdown}. The average total time to generate one chunk (49 frames) is 54.5 seconds. The primary computational bottleneck is the iterative denoising process of our spatial-temporal diffusion model (ST-Diff), which requires 42.5 seconds for 50 steps, accounting for approximately 78\% of the total time. In contrast, all 3D-related components are highly efficient: estimating the initial 3D representation with TTT3R takes 5.8s, optimizing the 3DGS cache takes only 2.5s, and forward warping is near-instant at 0.2s. This analysis demonstrates that the 3D-aware caching and conditioning, while critical for quality and consistency, add only a minimal computational overhead (8.5s total) compared to the main generative backbone.
\section{Conclusion}
\label{sec:conclusion}
In this work, we propose WorldWarp, a novel autoregressive framework for long-range, geometrically-consistent novel view extrapolation. Our method is designed to overcome the key limitation of prior work: the inability of standard generative models to handle imperfect 3D-warped priors. We introduce the ST-Diff model, a non-causal diffusion model trained with a spatially-temporally-varying noise schedule. This design explicitly trains the model to solve the fill-and-revise problem, simultaneously filling blank regions from pure noise while revising distorted content from a partially-noised state. By coupling this model with an online 3D geometric cache to avoid irreversible error propagation, WorldWarp achieves state-of-the-art performance, setting a new bar for long-range, camera-controlled video generation.

{
    \small
    \bibliographystyle{ieeenat_fullname}
    \bibliography{main}
}

\clearpage
\setcounter{page}{1}
\maketitlesupplementary

\begin{figure*}[t]
    \centering 
    \includegraphics[width=0.99\textwidth]{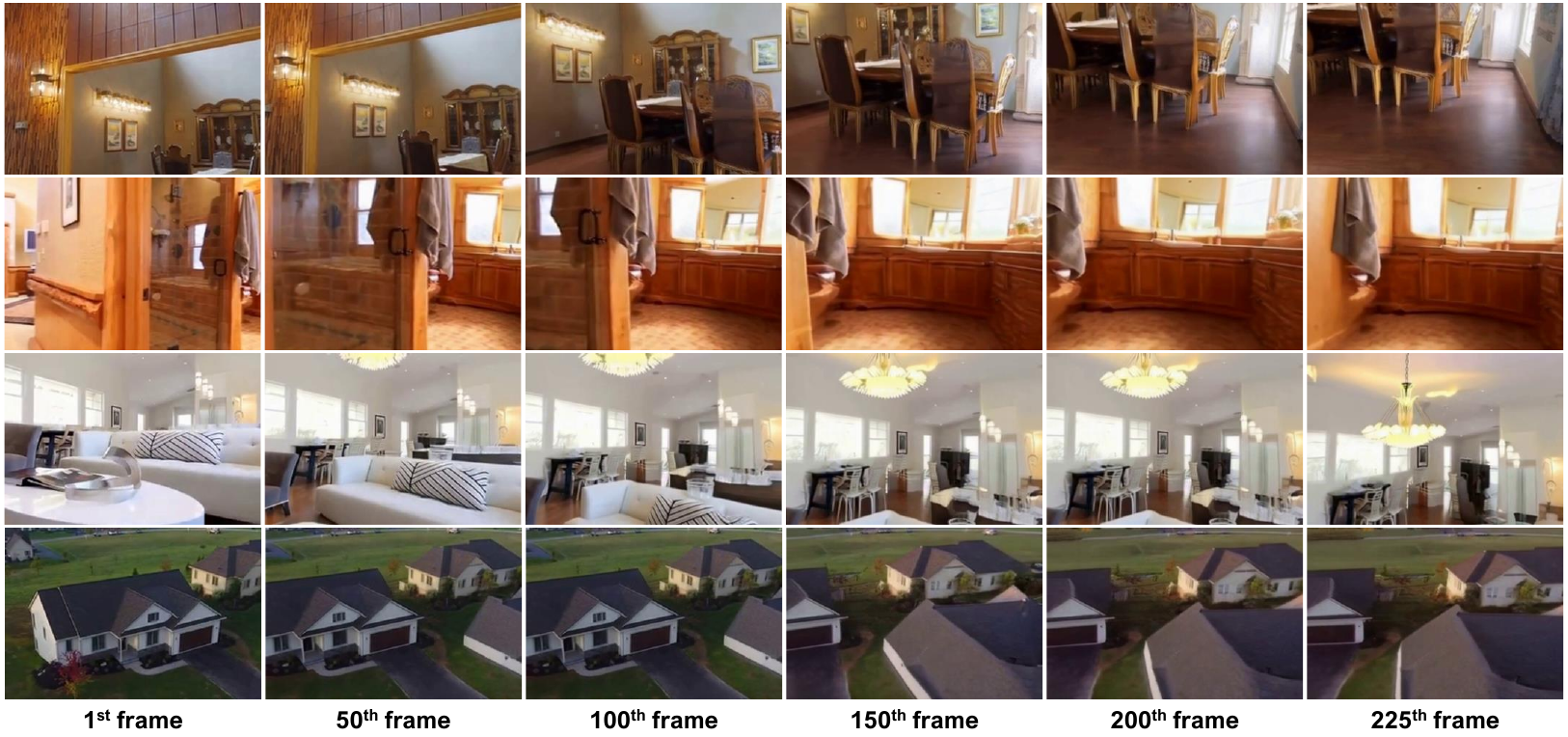}
      \captionsetup{font=small}
    \caption{Qualitative results on the RealEstate10K~\cite{re10k} datasets.} 

    \label{fig:supp_re10k}
\end{figure*}

\section{Implementation Details}

\noindent\textbf{Training.} 
We implement our WorldWarp using the Wan2.1-T2V-1.3B diffusion transformer~\cite{wan} as the generative backbone. All parameters of the model are fully fine-tuned in an end-to-end manner. The model is trained for 10k steps with a global batch size of 64 (8 per GPU) on 8 NVIDIA H200 GPUs. We utilize the AdamW optimizer with a learning rate of $5\times10^{-5}$ and apply a 1,000-step linear warmup. The training video resolution is set to $480 \times 720$.

\noindent\textbf{Inference.} 
The video generation process is initiated from a single source image. Subsequent content is synthesized auto-regressively in chunks of 49 frames. To ensure temporal continuity, we utilize a fixed context overlap of 5 frames for every iteration following the initial chunk. To establish the global coordinate system, we first estimate camera poses and intrinsics from the reference video using TTT3R~\cite{ttt3r}. For generation lengths exceeding the reference trajectory, we employ a velocity-based extrapolation strategy, computing linear velocity for translation and Spherical Linear Interpolation (SLERP) for rotation based on a 20-frame smoothing window. During the generation of each chunk, we optimize the online 3DGS cache for 500 iterations with a learning rate of $1.6 \times 10^{-3}$ to render the warped priors $\mathcal{X}_{s \to \mathcal{V}}$.

We utilize the spatially-temporally schedule described in \cref{fig:inference} in the main text for the reverse diffusion process. Specifically, the latent representations of the 5 context overlap frames are enforced as hard constraints. For the target frames, we set the strength parameter $\tau = 0.8$. Consequently, regions with valid geometric warps ($\mathbf{M}_{\text{latent}}=1$) are initialized with a reduced noise level $\sigma_{\text{start}}$ corresponding to $\tau$, preserving the structural integrity of the 3D cache. Conversely, occluded or blank regions ($\mathbf{M}_{\text{latent}}=0$) are initialized with standard Gaussian noise ($\sigma_{\text{filled}}=\sigma_{T_N} \approx 1.0$) to facilitate generative inpainting. The diffusion model employs a Flow Match Euler Discrete Scheduler with 50 denoising steps.

\section{More Experiment Results}

\subsection{Visualization Results}

\begin{figure*}[t]
    \centering 
    \includegraphics[width=0.99\textwidth]{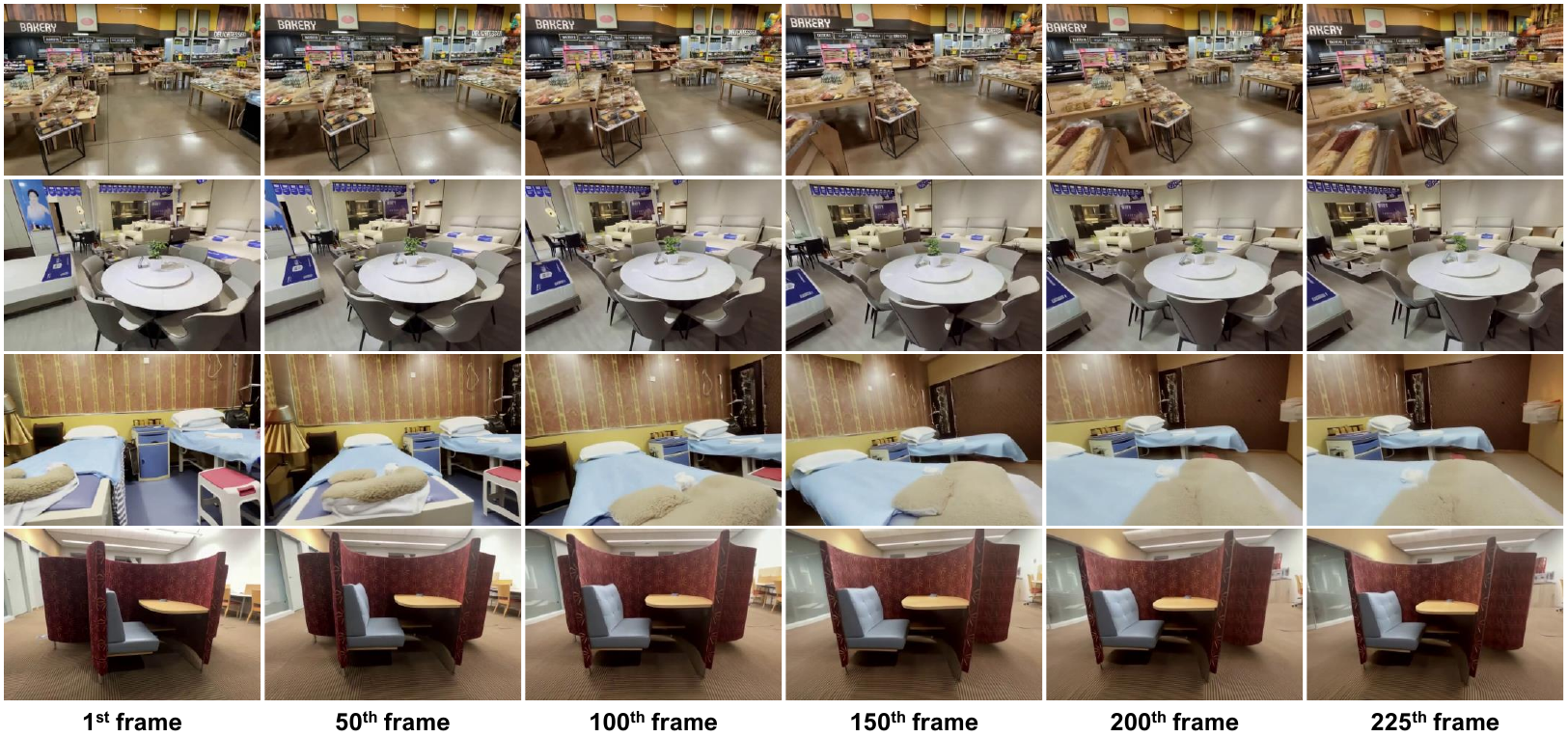}
      \captionsetup{font=small}
    \caption{Qualitative results on the DL3DV~\cite{dl3dv} datasets.} 
    \label{fig:supp_dl3dv}
\end{figure*}

We illustrate more results on the RealEstate10K~\cite{re10k} and DL3DV~\cite{dl3dv} datasets in \cref{fig:supp_re10k} and \cref{fig:supp_dl3dv}. Please refer to video supplementary material for more results.

\begin{figure*}[t]
    \centering 
    \includegraphics[width=0.99\textwidth]{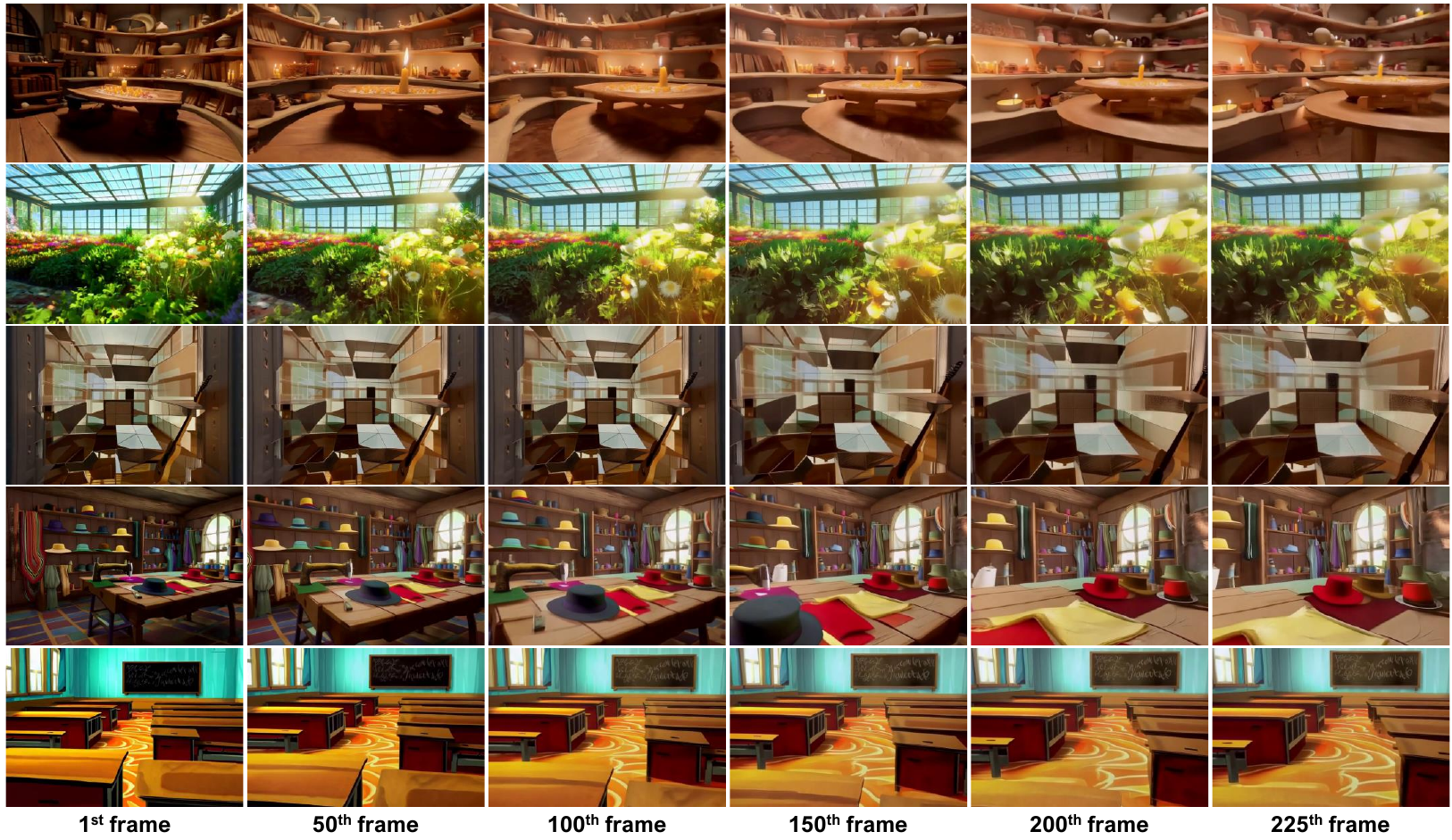}
      \captionsetup{font=small}
    \caption{\textbf{Generalization to Out-of-Distribution Artistic Styles.} 
We evaluate the robustness of ST-Diff by generating video sequences conditioned on diverse artistic text prompts (e.g., ``Van Gogh style'', ``Studio Ghibli style''). 
The visualized chunks demonstrate that our method successfully synthesizes high-quality stylized content while maintaining rigorous \textbf{3D geometric consistency} across the autoregressive generation. 
This confirms that our geometry-aware fine-tuning strategy effectively incorporates structural control without compromising the semantic and aesthetic generalization capabilities of the pre-trained diffusion backbone.}
    \label{fig:supp_ood}
\end{figure*}

\begin{figure*}[t]
    \centering 
    \includegraphics[width=0.99\textwidth]{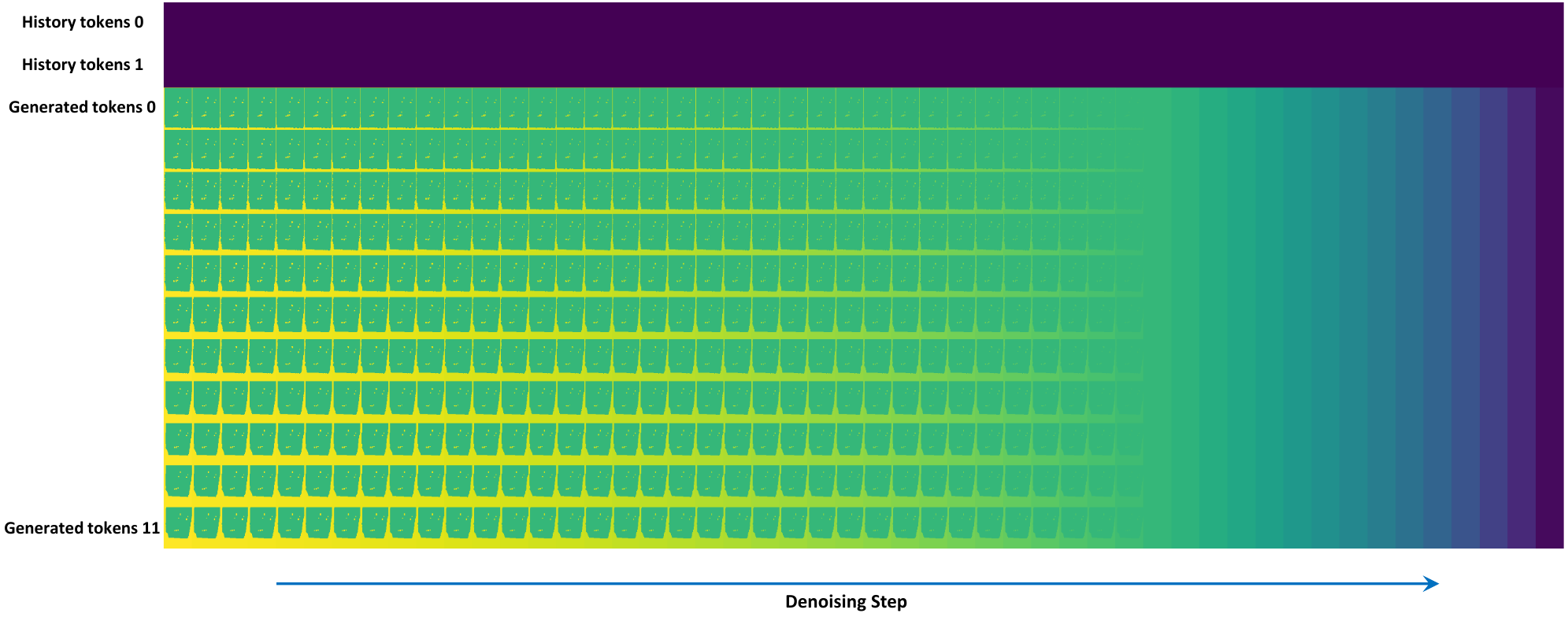}
      \captionsetup{font=small}
    \caption{\textbf{Visualization of the Spatially-Adaptive Noise Schedule.} 
    We visualize the schedule matrix $\mathbf{\Sigma}_{\mathcal{V}}$ across the reverse diffusion process. The horizontal axis represents the progression of denoising steps (from $T=999$ to $0$), while the vertical axis corresponds to the sequence of temporal tokens (13 tokens derived from 49 frames). The top two rows represent the \textbf{History Tokens} (context), which are enforced as hard constraints (dark purple, $\sigma=0$). The subsequent rows represent the \textbf{Generated Tokens}, where the noise levels are spatially modulated: (1) \textbf{Valid warped regions} are held at a reduced noise level $\tau$ (intermediate green) to preserve explicit geometry; and (2) \textbf{Occluded regions} are initialized with maximal noise (yellow, $\sigma \approx 1.0$) to enable the generative synthesis of novel content.}
    \label{fig:schedule}
\end{figure*}

To further demonstrate the robust generalization capacity of ST-Diff, we extend our evaluation beyond standard photorealistic benchmarks to scenes rendered in diverse artistic styles in \cref{fig:supp_ood}. By prompting the model with specific stylistic descriptors, such as "Van Gogh style" or "Studio Ghibli style," we generate a variety of stylized video sequences. The results illustrate that our method successfully synthesizes these highly stylized scenes while strictly preserving the underlying 3D geometric consistency. These qualitative results validate that our proposed training strategy effectively integrates fine-grained geometric control without sacrificing the rich semantic and aesthetic generalization capabilities inherent in the pre-trained model. This confirms that adapting the foundation model into an asynchronous diffusion framework does not compromise its ability to interpret open-domain text prompts.

\subsection{Analysis of Spatially-Adaptive Noise Dynamics}
To validate the effectiveness of our geometry-aware inference strategy, we visualize the evolution of the noise schedule matrix $\mathbf{\Sigma}_{\mathcal{V}}$ throughout the reverse diffusion process. As shown in \cref{fig:schedule}, the visualization is structured as a spatio-temporal grid where each row corresponds to a latent temporal token $t$ (derived from the 49 video frames via VAE encoding) and columns progress through the denoising steps from left ($T=999$) to right ($T=0$).

The map explicitly corroborates our dual-schedule formulation. The top two rows, representing the 5 history context frames (corresponding to $\sim$2 latent tokens), remain fully constrained with zero noise (dark purple) throughout the process, ensuring seamless transitions from previous chunks. In the subsequent 11 rows (the generated tokens), we observe a distinct spatial modulation. The valid geometric regions, projected from the 3DGS cache, are maintained at a reduced noise level $\tau$ (intermediate green/teal) to preserve high-fidelity structural details. In contrast, occluded or blank regions are initialized with high-variance noise (yellow) to facilitate the generative hallucination of new content. This confirms that our model effectively balances geometric preservation with generative inpainting during the autoregressive process.

\section{Limitations}
\label{sec:limitations}

Despite the effectiveness of ST-Diff in generating geometrically consistent long-term videos, our method is subject to certain limitations common to autoregressive video generation frameworks.

\paragraph{Error Accumulation in Long-horizon Generation.}
Although our model is trained in an asynchronous diffusion manner, where we apply varying noise strengths to different frames and spatial regions to mimic inference conditions, generating infinite-length video sequences with perfect fidelity remains an unresolved challenge. In our autoregressive pipeline, the generated output of one chunk serves as the historical context for the next. Consequently, minor visual artifacts or geometric inconsistencies can propagate and accumulate over time. For extremely long sequences, such as those exceeding 1000 frames, this drift can eventually lead to degradation in visual quality or geometric stability. This remains a persistent issue shared by state-of-the-art video generation methods.

\paragraph{Dependency on Geometric Priors.}
Our method operates on the premise that forward-warped images provide strong geometric \textit{hints} for generation. Therefore, our performance is heavily dependent on the accuracy of the upstream 3D geometry foundation models, such as TTT3R~\cite{ttt3r} or VGGT~\cite{vggt}, used for depth and camera pose estimation. In scenarios where these pre-trained estimators struggle, including complex outdoor environments with extreme lighting, transparency, or lack of texture, the estimated depth maps and poses may be inaccurate. This results in incorrect warping results that deviate significantly from the true geometry. While ST-Diff is designed to correct artifacts, it may fail to recover high-quality frames when the geometric guidance is fundamentally flawed or contains excessive noise.
\end{document}